\documentclass[runningheads]{llncs}
\usepackage[T1]{fontenc}
\usepackage{graphicx}
\usepackage{comment}

\usepackage{booktabs,array,colortbl}
\definecolor{manatee}{rgb}{0.59, 0.6, 0.67}
\definecolor{lightgray}{rgb}{0.83, 0.83, 0.83}

\begin{document}
\title{Computational-level Analysis of \\ Constraint Compliance for General Intelligence}

\titlerunning{Computational-level Analysis of Constraint Compliance for AGI}
\author{Robert E. Wray\orcidID{0000-0002-5311-8593} \and
Steven J. Jones\orcidID{0000-0003-1942-6354} \and
John E. Laird\orcidID{0000-0001-7446-3241}}
\authorrunning{R. Wray et al.}
\institute{The Center for Integrated Cognition \\ IQM Research Institute, Ann Arbor, MI 48105 USA \\
\url{URL: integratedcognition.ai} \\
\email{\{robert.wray,steven.jones,john.laird\}@cic.iqmri.org}
}
\maketitle              %
\begin{abstract}
Human behavior is conditioned by codes and norms that constrain action. Rules, ``manners,'' laws, and moral imperatives are examples of classes of constraints that govern human behavior. These systems of constraints are ``messy:'' individual constraints are often poorly defined, what constraints are relevant in a particular situation may be unknown or ambiguous, constraints interact and conflict with one another, and determining how to act within the bounds of the relevant constraints may be a significant challenge, especially when rapid decisions are needed.  General, artificially-intelligent agents must be able to navigate the messiness of systems of real-world constraints in order to behave predictability and reliably. In this paper, we characterize sources of complexity in constraint processing for general agents and describe a computational-level analysis for such \textit{constraint compliance}. We identify key algorithmic requirements based on the computational-level analysis and outline a limited, exploratory implementation of a general approach to constraint compliance.

\end{abstract}

\section{Introduction}

Rules, social norms (e.g., ``manners''), laws, and moral imperatives are examples of various classes of \textit{constraints} that govern human behavior. Systems of constraints are ``messy:'' individual constraints are often poorly defined; the constraints relevant in a particular situation may be unknown or ambiguous; constraints interact and conflict with one another; and determining how to act rapidly within the bounds of relevant constraints may itself be a significant challenge. Yet humans routinely and robustly overcome the messiness of conforming to many simultaneous and often ill-defined constraints. 

Notably, humans can also rapidly adapt their task performance to new constraints. A driver who has always driven on the left can, with just a little deliberation and practice, shift to driving on the right side of the road. A traveler has the ability to recognize and to adapt to overt local customs related to greetings, meals, etc. Humans can quickly and robustly adapt to novel constraints, even when those novel constraints interact with familiar constraints and tasks.

Today's AI systems, in contrast, generally elide or ignore the messiness of complying with real-world constraints. They often encode a designer's interpretation of constraints (e.g., by knowledge engineering or learning from a human-defined policy) and are designed for limited, pre-specified operating contexts~\cite{mani_artificial_2021_custom}. These systems conform to engineered constraints unfailingly but inflexibly. The encoding of constraints (along with designer assumptions) is tightly integrated with task specifications, making it difficult for the systems to adapt to new operating environments. For example, compare the relative immediacy of human adaptation to driving on their ``opposite'' side of the road for the first time vs. an autonomous driving system as trained today or the present limitations of large language models to conform to ethical guidance when producing responses \cite{weidinger_ethical_2021_local}.

These approaches can be acceptable for narrow AI but, as human intelligence suggests, a general artificial intelligence requires an ability to reason about its constraints (and conflicts), resolve ambiguity, determine how it should proceed given awareness of constraints, and be rapidly adaptive to new constraints. We introduce a broader approach to constraints, \textit{constraint compliance}, intended to provide an agent with the capacity to comply with real-world constraints.

We consider the computational requirements for this more comprehensive approach to compliance to systems of constraints, emphasizing general intelligence. That is, we seek to identify a computational approach that is constraint-compliant, domain general (not specific to an application or a task domain), and robust to the complexities that ``real world constraints'' introduce. We outline sources of ``messiness'' relevant to constraint processing and present a computational analysis of an overall constraint-compliance process, enumerating five distinct types of processing steps the agent must make. We then outline an initial algorithmic-level exploration of a constraint-compliance process. Finally, based on the analysis and exploratory implementation, we identify four algorithmic challenges that require additional analysis and research in order to realize comprehensive constraint compliance.

\section{Sources of Complexity in Constraint Processing}
Here, we enumerate specific sources of complexity and challenge for comprehensive constraint compliance. This ``messiness'' derives from many sources spanning the environment, the agent's task(s), its internal capabilities and assumptions, and the specification of constraints themselves. 

We illustrate using examples from Sudoku-puzzle solving and automobile driving. Sudoku is a canonical constraint satisfaction problem (CSP) \cite{lynce_sudoku_2006} and offers an effective contrast between classical constraint satisfaction 
\cite{dechter_constraint_2003} and the more comprehensive account of constraint processing we examine here.\footnote{More recent approaches to constraints extend the coverage of classical approaches but do not span all the forms of messiness  we consider \cite{rossi_building_2019_local}.}

Automobile driving offers a specific, familiar domain in which the real-world challenges of comprehensively complying to constraints arise; constraints abound in driving. This choice of domain is illustrative only: our goal is to develop a general approach to constraint compliance, not one specific to a single domain. 

\subsection{Partial Observability}
\label{section:observability}

In Sudoku, the puzzle state is fully available. The rules of the game (the constraints) can be readily applied after each move. Agents in real-world environments cannot generally sense everything and their actions often have uncertain outcomes. While partial observability and uncertainty have broad implications for agent reasoning \cite{pearl_reasoning_1990}, they impose specific demands for constraint compliance.

As one example, student drivers in the US are taught to ``maintain a 3-second distance when following on dry payment.'' Unlike a speed limit, where a speedometer provides an immediate gauge of one's speed, fully complying with this constraint requires that the driver visually attend to and continually assess the distance and current speed of their vehicle vs. the one in front and adjust speed to maintain the minimum distance.\footnote{Some newer cars offer an indicator for travelling too closely. Thus, with a different embodiment, this constraint no longer requires active measurement.}
Because not all constraint-relevant parameters are directly accessible to the agent, the agent must take action to determine the compliance of its behavior with that constraint.

\subsection{Dynamic, Fail-hard Environments}
\label{section:dynamic}

Dynamic environments compound the sources of messiness. Generally, dynamics amplifies the need for satisficing algorithmic solutions \cite{simon_models_1957,gigerenzer_fast_2004_custom}. Algorithms must (minimally) be responsive to the dynamics of the environment. A driver cut off in traffic by another car cannot pause to reason about all the potential instantiations and implications of its constraints in this unexpected situation, it must continue to drive and manage its constraint compliance over time. The specification of constraints themselves can also change due to environment dynamics (e.g., new traffic laws). Finally, interactions between constraints (below) may only become evident as a dynamic situation unfolds.

\subsection{Abstract and Poorly-defined Constraints}
\label{section:abstract-poor}

Real-world constraints are often ambiguous, abstract, and/or incomplete in their definition, giving rise to the challenge of interpreting and operationalizing such constraints \cite{wray_incorporating_2021_custom}.  In puzzles like Sudoku, however, constraint definition is unambiguous. Terms (cell, column, row) have immediate and direct correspondence to the representation of the puzzle. The constraints (or rules) defining the puzzle are also unambiguous.  

In contrast, many constraints in driving are abstract (``drive defensively'') or ambiguous (``do not follow too closely''). Terms used in constraints require a mapping onto one's internal representation that is not always consistent from person to person. ``Use caution near pedestrians'' depends on how one understands and applies both ``caution'' and ``near'' and perhaps also ``pedestrian.'' 

It may seem possible to overcome this source of messiness by directly encoding the ``meaning'' of constraints into an agent. However, resilience and robustness in open-ended environments requires disintermediation of the encoding and interpretation of constraints. Attempting to specify in advance how the agent should interpret constraints in every situation is likely to fail when the agent (inevitably) encounters a situation not anticipated by a system designer.

\subsection{Implicit Context Specification}
\label{section:implicit}

The definitions of real-world constraints often imply additional parameters or conditions rather than explicitly defining them. Most importantly, constraint specifications typically omit the context(s) in which they should apply.   By ``context,'' we mean a set of situations that share common, salient features.  The ``automobile driving'' context includes cars, roads, traffic laws, traffic signals, etc. Similar but different contexts can have constraints that prescribe very different behaviors. For instance, ``do not pass on the right'' is a constraint relevant in countries where vehicles are driven on the right side of the road, but is not apt (most of the time) for countries where vehicles are driven on the left. 

For Sudoku, there is an implicit but  single context. Thus implicit specification poses no problem to the classical approach to constraints. 

\subsection{Interactions \& Conflicts among Constraints, Tasks, \& Contexts} 
\label{section:interactions}

Interactions and conflicts among constraints and between task(s) and constraints can arise frequently. An accident or road construction causes re-routing of traffic into normally oncoming traffic lanes. A text message notification draws attention when attending to the road is required (sometimes by law). To what extent should one obey traffic laws when transporting someone in dire medical distress? The specific instantiation of constraints grounded within a given situation will indicate competing and sometimes conflicting choices for the agent.

The design of Sudoku ensures that constraints are collectively coherent (simplified by the single context). Generally, classical approaches to satisfying constraints only provide solutions when sets of constraints are coherent, obviating conflicts. More recent approaches support ``soft constraints'' \cite{meseguer2006soft} which support prioritization of constraints when conflicts arise; the overall set of constraints remains coherent when prioritization is taken into account.

In real-world situations, conflicts cannot always be resolved via \textit{a priori} prioritization; an agent must sometimes knowingly violate a constraint. If a car is cut off in heavy traffic, it is probably more important to maintain speed and slowly build distance between the car ahead than to sharply brake in order to regain compliance with the following-distance constraint. From the point of view of constraint compliance, the agent is often likely to be in situations, imposed by dynamics directly but also sometimes at its choosing given the dynamics, to violate some constraints and to repair violations as the evolving situation allows.

\section{Computational-level Analysis}

We now present a computational-level account of the information processing tasks necessary for constraint compliance given the many sources of ``messiness'' above.  A computational-level analysis emphasizes \textit{what} steps are required to achieve constraint compliance and identifies requirements for \textit{how} the capability may be realized at the processing and representation (``algorithmic'') level \cite{marr_vision_1982}.

\subsection{Functional Role}
\label{section:functional}

\begin{figure*}[t]
    \centering
    \includegraphics[width=0.98\textwidth]{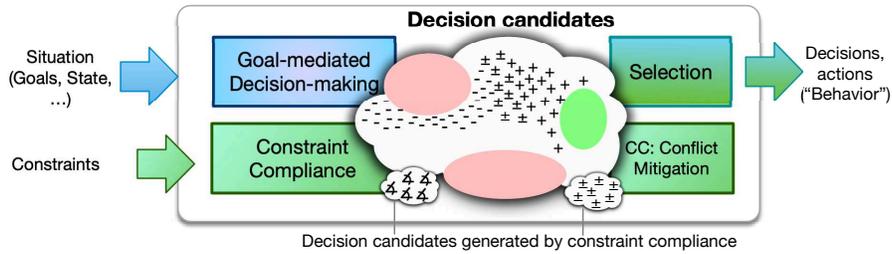}
    \caption{At the computational level, the purpose of constraint compliance is to ensure that decision making takes constraints relevant to the agent's situation into account.}
    \label{fig:computatational-level}
\end{figure*}

The functional role of constraint compliance is to modulate agent decisions (and thus behavior) so that constraints relevant to the current situation inform agent choices. In Figure~\ref{fig:computatational-level}, the agent's goal-focused decision process (blue) generates candidate choices and selects among them. The primary input to this decision process is the current situation (including environment state, external goals, history, etc.) and the output is a decision. A decision could be a commitment to a long-term course of action (e.g., a plan), an intermediate subgoal, or an immediate action. Over time, the sequence of decisions produces behavior (e.g., ``driving''). We illustrate constraint compliance (green) parallel to the goal-mediated decision process of the agent and external constraints as a distinct input. This separation is for illustration only; at the algorithmic level, solutions may integrate constraint-compliance with goal-mediated decision processes.

As suggested by the figure, the agent commits to its decisions from a (potentially very large) space of candidate choices. At the computational level, we do not assume that the agent has an explicit representation of this space; in the figure, the cloud represents a conceptual space from which a specific decision might be drawn. For example, the decision process might choose actions based on a learned policy, where the space is implicit in mappings from states to actions.  

Functionally, constraint compliance augments candidate choices produced by the goal-mediated decision process by indicating the acceptability/desirability of the candidates with respect to relevant constraints. The figure shows parts of the candidate space that are required (green), prohibited (red), desirable (+), and undesirable (-) choices. Because constraints can conflict (\S\ref{section:interactions}), some candidates are labeled as both desired and undesired ($\pm$); however, conflicts can occur in any combination.
The selection process (green/blue) now evaluates the candidates and the desirability of those candidates.

Constraint compliance can also add new candidates. The grounding process can suggest candidates to take new actions (e.g., measurements to evaluate individual constraints; \S\ref{section:observability}). In order to mitigate conflicts in constraints, the selection process may produce new candidates as well. Thus, in contrast to classical constraint satisfaction (where the application of constraints reduces choices), constraint compliance can produce additional choices. It also enables the agent to choose courses of action that are not necessarily consistent with all constraints.

\subsection{Processing Steps for Constraint Compliance}
\label{section:process}
\begin{figure*}[t]
    \centering
    \includegraphics[width=0.98\textwidth]{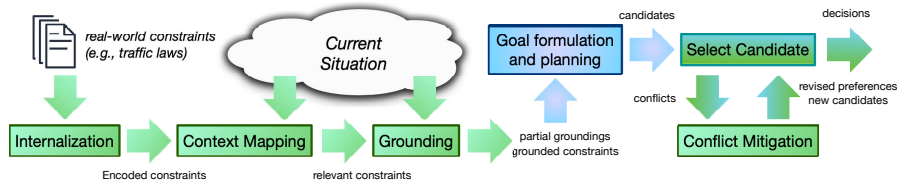}
    \caption{Simple process model for constraint compliance.}
    \label{fig:lifecycle}
\end{figure*}

What computational tasks are performed by the constraint-compliance process?  
Figure~\ref{fig:lifecycle} illustrates a high-level process. The specific sequence of steps illustrates both a simple process model and how we are exploring constraint compliance at the algorithmic level and integrating it with decision making (see \S\ref{section:algorithmic-level}).

The agent's internal representations of constraints derive from real-world constraints defined externally (e.g., a law). \textit{Internalization} results in encoding of constraints in agent memory. Next, \textit{Context Mapping} compares encoded constraints to the current situation, identifying what constraints are (potentially) relevant in a given situation. Context mapping results in a set of situation-relevant but abstract (not grounded) constraints.  

\textit{Grounding} then maps abstract constraints to specific objects in the environment. In our explorations to date, both complete and partially-grounded constraints are re-represented as goals in order to  exploit an existing agent's planning capability (\S\ref{section:algorithmic-level}). Planning generates candidates for \textit{Selection} which is now extended with an ability to assess the acceptability of decision candidates based on the constraints. When conflicts arise, selection is augmented with \textit{Conflict Mitigation}, which may lead to the generation of of alternative courses of action. Below, we further describe these steps, focusing especially on how ``messiness'' motivates and/or introduces additional requirements for individual steps.

\textbf{Internalize constraints.}
Real-world constraints (typically) are defined external to the agent. Thus, an initial step in constraint compliance is to interpret the external constraint; that is, to map the  external representation of the constraint to concepts as represented within the agent. 

Abstract and poorly-defined constraints (\S\ref{section:abstract-poor}) introduce challenges to simple encoding. The agent may not possess internal representations that align with the conditions in the external constraint and thus algorithmic approaches to internalization will entail methods that allow an agent to assess mappings between external conditions and internal representations.

\textbf{Identify situational context(s).}
Conforming to real-world constraints requires an agent to recognize which constraints are relevant to its situation. However, the applicable situation (or general characterization of situations: contexts) are often implicit in the specification of constraints (\S\ref{section:implicit}). An agent can often learn associations between contexts and constraints through experience (which can include instruction) but a core challenge is that constraint specifications themselves do not (usually) specify applicable contexts.  

A second challenge results when the composition of contexts interact in ways that make previously learned mappings inapt or invalid (\S\ref{section:interactions}).
Anticipating and evaluating all possible compositions of all possible contexts is not feasible. Thus, general intelligence requires the capacity to consider and to evaluate constraints in novel contexts as behavior is being generated. 

Context recognition itself is a challenge \cite{gershman2017context}. For an algorithmic implementation of constraint compliance, all that is needed is that the agent recognize ``this constraint is relevant in my current situation.'' However, a full solution to constraint compliance appears to require context recognition processes as well.

\textbf{Instantiate constraints in a situation (Grounding).}  
As an agent behaves in its environment, it must determine how constraints might apply in its current situation. Grounding is distinct from internalization and context identification; it requires that the agent shift from general consideration of a constraint to determining if/how it should be instantiated in the agent's current environment. 

Grounding is often straightforward. However, partial observability (\S\ref{section:observability}) and abstract constraints (\S\ref{section:implicit}) can require a search over potential instantiations of a constraint, rather than an immediate mapping. Thus, as constraints are expressed more abstractly and generally, the computational demand on the agent to determine \textit{how} that constraint may apply in the current situation increases. When new information is needed to complete grounding (e.g., a measurement as in  \S\ref{section:observability}), new candidate choices should be generated (\S\ref{section:functional}).

An agent's embodiment may lack an ability to directly observe features needed to instantiate a constraint. Nonetheless, the agent should still attempt to respect applicable constraints. Thus, grounding requires prospective instantiation with incomplete information.

\textbf{Integrate constraints in decision-making (Selection).}
At a minimum, the agent's selection process must take into account both agent goals and constraints for constraint compliance. When the set of applicable constraints are fully grounded and present no conflicts, the selection process is straightforward.

Conflicts (below) and partial grounding complicate selection. The selection process must be sensitive to both taking action to find an instantiation for a partially grounded constraint and also the potential costs and risks associated with that search. Defining algorithmic approaches to selection in the presence of partial grounding is a significant novel challenge. 

\textbf{Identify and mitigate conflicts.}
When there are conflicts in the acceptability and desirability of candidate choices, the agent must either 1) choose one of the options given the conflicting choices or 2) attempt to identify new choices that resolve or mitigate the conflicts. Specific strategies could include attention/inattention (ignoring some constraints), prioritization of constraints, and replanning. A primary algorithmic-level challenge is to resolve and mitigate conflicts rapidly, given bounded rationality in a dynamic environment (\S\ref{section:dynamic}).

\section{Exploratory Algorithmic-level Prototype} 
\label{section:algorithmic-level}
In parallel with the top-down computational-level analysis, we have begun bot\-tom-up prototyping as well, focusing to date on algorithmic approaches to grounding, selection, and conflict mitigation. We use Soar  \cite{laird_soar_2012} as the target implementation level. Soar both constrains and informs definition at the algorithmic level. We introduce further design constraint at the algorithmic level by building on an existing agent designed to interactively learn tasks \cite{kirk_learning_2019_custom,mininger_expanding_2021}. The prototype is compatible with this agent's \textit{a priori} capabilities for interpreting language, planning task actions, executing plans, and learning from human instruction. 

\textbf{Grounding:} The prototype builds on language grounding already part of the agent, which can learn recognition structures for abstract goal specifications \cite{kirk_learning_2019_custom} and maintain consistent grounding across perceptual changes \cite{mininger_expanding_2021}. The primary focus is to explore how to support partial grounding of constraints. The agent can now indicate that some actions are desirable (in Selection; see below) because they lead to further information that could potentially complete the grounding. In this way, the agent is biased towards choosing candidates that lead to measurement actions, as suggested in Figure~\ref{fig:computatational-level}. 

\textbf{Selection:} The original agent uses an explicit goal representation to determine what to do next (typically via search-based planning, although it can ask for help from an instructor as well). In our initial implementation, as shown in Figure~\ref{fig:lifecycle}, we integrated constraint-compliance with selection by having the agent represent grounded constraints as goals (e.g., a speed limit constraint would be represented as a goal for speed to be less than the limit). 
This approach leverages the agent's planning capability. Candidate evaluations (from grounding) are implemented as Soar preferences for selecting plans, which maps selection directly onto an implementation/architecture-level capability of Soar. In the absence of conflicts (below), planning provides a solution that satisfies the (grounded) constraints, measurement actions (from partial groundings), and task actions.  

\textbf{Conflict Mitigation:} Consider two conflicting constraints relevant to driving in a medical emergency. The lawful speed limit and a general directive to preserve human life apply. These constraints can result in a conflict over the desired speed. 
Because plan choices are mapped onto Soar preferences, Soar responds to conflicting preferences with an impasse, a conflict detection system already part of Soar. Thus, we have also mapped the trigger for conflict mitigation onto an implementation-level process. Generally, resolving conflicts requires additional knowledge (e.g., in this case, some sense that preserving life is more important than respecting the speed limit) which can include various ways to include values in assessing choices \cite{arkin_moral_2011,giancola_ethical_2020}.

\section{Discussion and Implications} 
While limited and preliminary, the initial prototype highlights examples of representation and process (algorithmic-level choices) and how these choices may interact with the implementation level. 
We now consider implications for future work at the algorithmic level to realize general constraint compliance.

\textbf{Online, Incremental Learning:} For an AGI, the set of contexts and constraints is potentially huge, it is infeasible to prepare for every contingency, and dynamics often demands rapid response. Together, these conditions point toward algorithmic solutions that employ online, incremental learning. This implication mirrors human learning and is consistent with the transition from more deliberate and explicit (System 2) to more implicit and automatic (System 1) reasoning \cite{kahneman_thinking_2011}. However, it contrasts with recent approaches that emphasize pre-training to ensure conformance to various operational and safety constraints \cite{garcia_comprehensive_2015}. 

\textbf{Senses of Familiarity, Novelty, and Surprise:} Familiarity, novelty, and surprise are important signals in human (and animal) regulation of behavior \cite{barto_novelty_2013_custom}. Realizations of familiarity, novelty and surprise may be useful for meta-cognitive regulation of constraint compliance in task performance. An open question is whether a sense of familiarity (and other signals) are best realized in the implementation level (e.g., extension to Soar) or at the algorithmic level.

\textbf{Anticipation based on Partial Information:} Near-term anticipation of future states is central to functional and neurological accounts of human intelligence \cite{bubic_prediction_2010}. Humans readily anticipate the potential impact of constraints on behavior and adapt behavior in advance of a potential constraint violation. 
 Our exploration identified a need for anticipation in grounding. An agent needs  strategies to decide which  potential groundings to attend to, given many potential groundings (with many implications). Indicators of potential threats to constraint compliance would provide a coarse attention mechanism to bias grounding processes toward more important constraints.

\textbf{Domain Knowledge:} Choosing to prioritize some constraints over others requires general knowledge of the world. Having such knowledge may be as important to the results of constraint compliance as the algorithms that realize its functions. This dilemma points to one of the rationales for adopting an agent that can learn from instruction. Because research agents will often lack knowledge, our agent can actively seek input to gain missing knowledge about conflicts. While this does not resolve the dependence of constraint compliance on general knowledge, it does provide a means to explore algorithmic realizations in a way that makes the required domain knowledge explicit and transparent.

\subsubsection{Acknowledgments}

This work was supported by the Office of Naval Research, contract N00014-22-1-2358. The views and conclusions contained in this document are those of the authors and should not be interpreted as representing the official policies, either expressed or implied, of the Department of Defense or Office of Naval Research. The U.S. Government is authorized to reproduce and distribute reprints for Government purposes notwithstanding any copyright notation hereon.  We thank the anonymous reviewers for substantive comments and suggestions.

\bibliographystyle{splncs-suppressDOIURLISBN}
\bibliography{zotero-transoar,bob-bibliography,agi23-local}
\end{document}